\documentclass[10pt]{article} 
\usepackage[accepted]{tmlr}

\usepackage{amsmath,amsfonts,bm}

\def\eqref#1{equation~\ref{#1}}

\def\1{\bm{1}}

\DeclareMathAlphabet{\mathsfit}{\encodingdefault}{\sfdefault}{m}{sl}
\SetMathAlphabet{\mathsfit}{bold}{\encodingdefault}{\sfdefault}{bx}{n}

\usepackage{hyperref}
\usepackage{url}

\usepackage{amsmath}
\usepackage{graphicx}
\usepackage{soul}
\usepackage{fancyhdr}
\usepackage{xcolor}
\usepackage{float}
\usepackage{graphicx}
\usepackage{bbm}
\usepackage{dblfloatfix}
\usepackage{booktabs}
\usepackage{enumitem}
\title{Language Models Are Better Than Humans at Next-token Prediction}

\author{\name Buck Shlegeris\thanks{Equal contribution} \email buck@rdwrs.com \\
      \addr Redwood Research
      \AND\name Fabien Roger$^*$ \email fabien.d.roger@gmail.com \\
      \addr Redwood Research
      \AND\name Lawrence Chan\thanks{Work primarily carried out while at Redwood Research.} \email lawrence.chan@live.cn \\
      \addr METR
      \AND\name Euan McLean \email emc031@gmail.com \\
      \addr FAR AI}

\def\month{07}  
\def\year{2024} 
\def\openreview{\url{https://openreview.net/forum?id=RNsnSLdmV7}} 

\begin{document}

\maketitle

\begin{abstract}
Current language models are considered to have sub-human capabilities at natural language tasks like question-answering or writing code. However, causal language models are not trained to perform well at these tasks; they are trained to accurately predict the next token given previous tokens in tokenized text. It is not clear whether language models are better or worse than humans at next-token prediction. To try to answer this question, we performed two distinct experiments to directly compare humans and language models on this front: one measuring top-1 accuracy and the other measuring perplexity on OpenWebText. In both experiments, we find humans to be consistently \emph{worse} than relatively small language models like GPT-Neo-1.3B or GPT-2-large at next-token prediction.\footnote{
Code is available at \url{https://github.com/FabienRoger/lm-game-analysis-main}.}
\end{abstract}

\section{Introduction}

Recent language models (LMs) have demonstrated impressive capabilities in natural language tasks, like writing convincing human-like text, coding, or answering general knowledge questions. However, LMs are not considered to have yet surpassed human performance at these tasks. But performance at such tasks is not a fair way of comparing LMs and humans. LMs are not explicitly trained to perform well at natural language tasks. Their loss function is simply \emph{next-token prediction}: accurately predicting the next token given previous tokens in tokenized text.

How good are modern LMs compared to humans at next-token prediction? While one can construct tasks in which humans make next-token predictions better than any language model, there have been no ``apples-to-apples'' comparisons on non-handcrafted datasets. To answer this question, we performed two experiments that directly compare humans to language models on next-token prediction, using the OpenWebText dataset \citep{openwebtext}.

Two natural ways of measuring the quality of next-token prediction are \emph{top-1 accuracy} and \emph{perplexity}. Top-1 accuracy is the fraction of times, over many predictions, that the predictor assigns the highest probability to the correct next token. This is relatively easy to measure, but does not capture information about the rest of the probability distribution that the predictor assigns over possible next tokens. A more all-encompassing but harder to measure variable is perplexity: defined by $2^{L}$ where the loss $L$ is the cross-entropy of the predictor's distribution and the true distribution over possible next tokens.

Contrary to some previous claims, we found humans to be consistently worse at next-token prediction than even small models like GPT-Neo-1.3B or GPT-2-large, in terms of both top-1 accuracy and perplexity. That is, even small LMs are superhuman at next-token prediction.

We structure this paper as follows. We first review claims made to date about the comparison between human and language model next-token prediction in Section \ref{sec:related_work}. In sections \ref{sec:human_top1accuracy} and \ref{sec:human_perplexity} we detail two small experiments we ran to measure human top-1 accuracy and perplexity respectively, along with their results. In section \ref{sec:conclusion} we discuss the implications of these results and conclude.

\section{Related work}
\label{sec:related_work}

One commonly cited source on the topic of human vs LM next-token prediction is a presentation by \citet{omohundro}. This presentation contains the claim that humans have a perplexity around 12, compared to the 20.5 of GPT-3 \citep{Brown2020LanguageMA}.

This comparison is problematic for two reasons, one small and one fatal. The smaller problem is a mismatch of text corpora: the language model statistics are word-level perplexities computed on Penn Tree Bank (PTB) \citep{penntreebank}, while the human word-level perplexity is estimated on the 1 Billion Words (1BW) benchmark \citep{Chelba2014OneBW}. This is a small problem in practice, as while GPT-3 was not evaluated on 1BW, GPT-2 performs slightly worse on 1BW than PTB. The bigger issue is the methodology used to estimate human perplexity. The value for human perplexity is quoted from  \citet{Shen2017EstimationOG}, in which, humans were asked to rate sentences on a 0-3 scale, where 0 means ``clearly inhuman'' and 3 means ``clearly human''. Then they computed a ``human judgment score'': the ratio of sentences rated 3 over those rated 0. They then fit a degree-3 polynomial regression to the LMs they had (of which the best was a small LSTM), which they extrapolated significantly out of distribution to acquire the ``human'' perplexity (see Figure 1 from \citet{Shen2017EstimationOG}). Due to this far extrapolation among other problems, we don’t agree with the claim that humans have perplexity 12.
\begin{figure}[!b]
    \centering
    \includegraphics[width=1.0\textwidth]{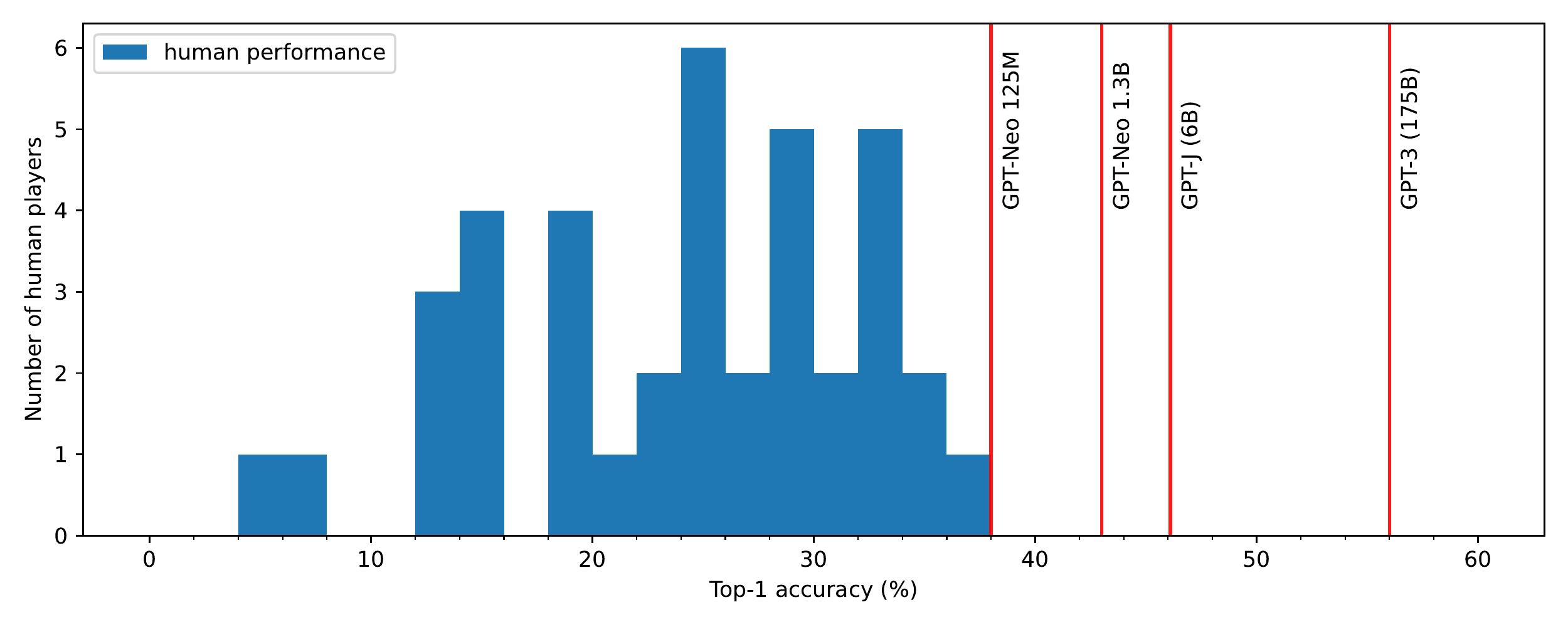}
    \vspace{-4mm}
    \caption{The distribution of human top-1 accuracy (how often a human guesses the correct next token given previous tokens) found in our study, with GPT-Neo-125M, GPT-Neo-1.3B, GPT-J (6B) and GPT-3 (175B) for comparison.}
    \label{fig:accuracy}
\end{figure}

Another claim is from OpenAI regarding the LAMBADA dataset \citep{openai2019}, where they give a perplexity of around 1-2 for humans (compared to 3 for 0-shot GPT-3). However, the authors don’t cite a source. The authors likely made an educated guess based on how the LAMBADA dataset was constructed. In addition, LAMBADA is a much restricted dataset, which consists of guessing single words requiring broad context. So this comparison isn’t very informative to the question of how good language models are at the task they’re trained on -- next-token prediction on typical internet text.

The most closely analogous study to ours is that of \citet{goldstein2022}. This found that an ensemble of 50 humans have a top-1 accuracy of 28\% vs 36\% for GPT-2, which is similar to what we saw for humans on webtext. However, they used a different, smaller dataset (the transcript from a podcast), which is not particularly representative of randomly-sampled English internet text.

\citet{Owens1997} pitted 8 humans against an n-gram model for missing word prediction. Humans achieved a top-1 accuracy of 26\%. This is related but not the same as top-1 accuracy for next-token prediction, since the prediction was on word-level rather than token level, and the predictions were conditioned on words appearing after the target word.

There are a number of more narrow datasets on which we know both human and LM performance, where some humans still outperform LMs. The MATH dataset \citep{Hendrycks2021MeasuringMP} is an example. But, to our knowledge, there are no apples-to-apples comparison between humans and modern LMs for webtext next-token prediction.

\section{Measuring human top-1 accuracy}
\label{sec:human_top1accuracy}

The main difficulty for comparing human and LM performance is that, unlike with language models, it is infeasible for humans to give their entire probability distribution for the next token in a sequence, since GPT-2 has a vocabulary size around 50,000.

One way to get around this is to simply measure top-1 accuracy by asking humans what token is most likely to come next. We can't derive human perplexity from this, but top-1 accuracy might still give us a reasonable measure of how well humans do at next-token prediction. According to this measure, humans are worse than all language models we tried, even humans who have practiced for more than an hour.

\subsection{Method}

The human participants were either staff or advisors of our lab, or members of the Bountied Rationality Facebook group. They were paid \$30/hour. There were 60 participants overall.

The language models we evaluated were GPT-Neo-125M, GPT-Neo-1.3B \citep{gpt-neo}, GPT-J-6B \citep{gpt-j}, and GPT-3 \citep{Brown2020LanguageMA}.

We set up a top-1 token prediction game website\footnote{Available at \url{http://rr-lm-game.herokuapp.com/}.}. We recommend playing the game if you want to get a sense of what next-token prediction is like.

In this game, the player makes their way through a random segment from the validation set of OpenWebText \citep{openwebtext}. At each token, they are asked to guess the single token they thought was most likely to come next. The correct next token is revealed to them, and they are once again asked to predict the most likely token to follow the revealed one. A total of 18530 guesses were made across participants.

Our website did not have any way for participants to guess ``visually empty'' tokens, such as newlines and half-Unicode characters (displayed as \texttt{<?>} on the website). We excluded cases where the correct guess was impossible from our analysis. They represent 6.3\% of questions asked to participants.

\subsection{Results}

The mean accuracy of the participants' answers on this top-1 task was 29\%. Of these players, 38 gave at least 50 answers, with an accuracy of 30\%. This accuracy is low compared to the accuracy of LMs: when measured on the same dataset, GPT-3 achieved an accuracy of 56\%. Even GPT-Neo-125M, achieved an accuracy above all players in our dataset. Figure \ref{fig:accuracy} shows the distribution of top-1 accuracies across participants, along with the accuracy of some language models.

We found that humans don't quickly get much higher performance with practice. There were 7 players who guessed over 500 tokens (taking roughly 5 hours), and achieved accuracies around the average of human performance (between 0.26 and 0.32).

\subsection{Additional analysis}

We study the effect of two possible confounders: the effect of overfitting of the comparison models and the effect of weird tokens on human performance.

GPT-Neo and GPT-J models were trained on the pile, which may contain some of the OpenWebText data used to compute the model accuracy. We study this effect by measuring the top-1 accuracy of GPT-Neo and GPT-J models on the first 256 tokens of 1024 randomly selected passages of the Pile \citep{gao2020pile} (train and validation set). We find that overfitting is very unlikely to be significant, as the top-1 accuracy on the train and val set of the pile are very close, as shown in Table \ref{table:acc-and-perplexity}.

\begin{table}[t]
\centering
\begin{tabular}{llll}
\toprule
Model & Split & Perplexity (per token) & Top-1 Accuracy \\
\midrule
GPT-Neo-125M & Validation & $29.26 \pm 11.22$ & $0.483 \pm 0.009$ \\ 
GPT-Neo-125M & Train & $22.76 \pm 1.62$ & $0.481 \pm 0.009$ \\ 
GPT-Neo-1.3B & Validation & $17.84 \pm 7.79$ & $0.547 \pm 0.009$ \\ 
GPT-Neo-1.3B & Train & $12.89 \pm 0.91$ & $0.545 \pm 0.009$ \\ 
GPT-J-6B & Validation & $12.49 \pm 4.42$ & $0.580 \pm 0.009$ \\ 
GPT-J-6B & Train & $9.57 \pm 0.64$ & $0.579 \pm 0.009$ \\
\bottomrule
\end{tabular}
\caption{Model accuracy and perplexity on the train and validation datasets of the Pile. We show 2-sigma standard deviations (over sequences) divided by $\sqrt{N_\text{sequences}}$ as uncertainties.}
\label{table:acc-and-perplexity}
\end{table}
\begin{figure}[!b]
    \centering
    \includegraphics[width=1.0\textwidth]{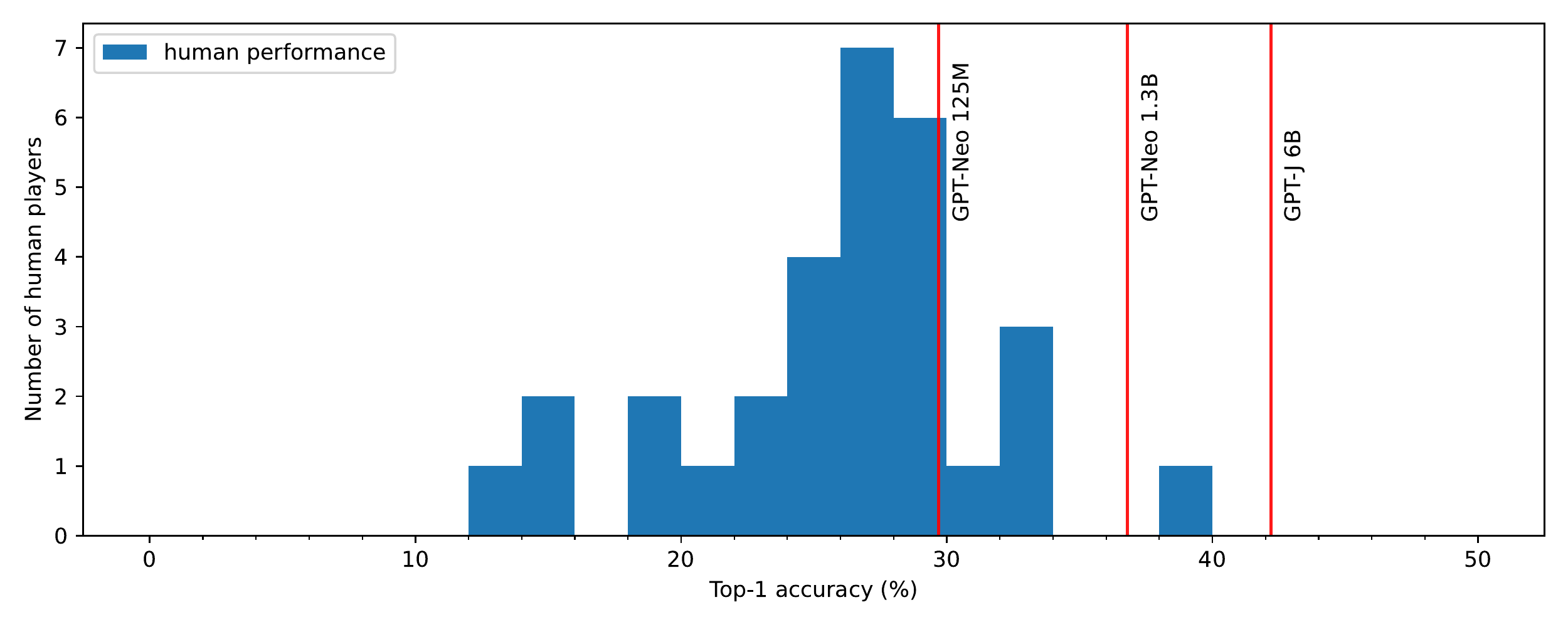}
    \vspace{-4mm}
    \caption{The distribution of human top-1 accuracy on a \textbf{filtered dataset} made out of single-word tokens, with GPT-Neo and GPT-J models of varying sizes for comparison.}
    \label{fig:accuracy-2}
\end{figure}

Human participants did not extensively study the tokenization scheme of GPT-2. They were helped by the website that prevented them from providing incorrect tokens, but tokenization difficulties may still have hindered their performance. To study the impact of this effect, we report in Figure \ref{fig:accuracy-2} the top-1 accuracy of models and human participants for tokens that correspond to words: we keep tokens that start with a space, that otherwise have only alphabetical characters in them, and followed by a token starting with a space, which represent 62\% of tokens from the dataset. We find that this reduces the gap between human participants and models considerably. It is unclear to what extent this is due to tokenization difficulties or to another bias introduced by filtering. For example, these numbers may reflect a bias humans may have towards single-word tokens, or LLMs' wide knowledge (relative to humans) about the frequency of rare proper nouns - which are often excluded by our filter.

\section{Measuring human perplexity}
\label{sec:human_perplexity}

Language models aren’t trained to guess the single most likely next token, so the previous task isn’t directly assessing language models on the task they’re trained on. They are trained to give a full distribution over next tokens, aiming to minimize their perplexity. Therefore, measuring top-1 accuracy may underestimate how good language models are relative to humans. In this section, we therefore try to compare human and language models' perplexity.

It is easy to measure models' perplexity, because language models produce a probability distribution for every next token. Unfortunately, humans aren’t able to quickly provide a probability distribution over the 50,000 possible tokens. So human perplexity must be estimated indirectly.

\subsection{Method}
\label{sec:perplexity_method}

To gather the relevant data, we set up a second language modeling game\footnote{Available at \url{https://rr-lm-game.herokuapp.com/whichonescored}} website. 54 humans participated. They were again either staff of our lab or members of the Bountied Rationality Facebook group, paid \$15 for answering a set of 120 rounds (taking roughly 30 minutes overall per participant). 19 participants answered all 40 questions of the first set, and 11 participants answered all 80 questions of the second set, answering a total of 1640 comparisons.

On each round, a prompt $c$ from the validation dataset of OpenWebText \citep{openwebtext} is shown (with a maximum length of 120 tokens). The participant is then asked to guess the likelihood of one candidate token $x$ being correct, given that one of the two candidates $x$,$y$ are correct. From this the ratio of probabilities of $x$ and $y$ can be determined, which we call $r(x,y|c)$.

From the responses we build a dataset of many values of $r(x,y|c)$ for many $x,y$ pairs. This data can be used to estimate human perplexity. We make the following assumption: for a given prompt $c$, all humans have the same probability distribution over next tokens $h(y|c)$. In this case, the responses from participants can be written as $r(x,y|c) = \frac {h(x|c)}{h(y|c)}$.

In the following sections we detail how we translated the dataset into an estimate of human perplexity $P_h$.

We used importance sampling to generate the candidate tokens (see Section \ref{sec:importance_sampling}). This way of sampling can introduce a bias to the data, so we modified the game slightly to correct for this bias (Section \ref{sec:sample_bias}). In Section \ref{sec:uncertainties} we describe how we estimate an uncertainty for the results.

\subsubsection{Estimating perplexity from a few relative probabilities with importance sampling}
\label{sec:importance_sampling}

Let $T$ be the true distribution of tokens $y$ after a context $c$. Human perplexity $P_h$ is defined as
\begin{align}
    \label{eq:human_loss_def}
    P_h = 2^{L_h},\quad L_h =-\mathbbm{E}_{(C,Y)\sim T}[\log h(Y|C)].
\end{align}
$L_h$ is referred to as the human loss function.
We can’t directly ask a human for the probability of the true token $h(y|c)$ without spoiling them on the answer, and it would be cumbersome to ask for their whole probability distribution. We can do better by asking for relative likelihoods. For a given context $c$ and true token $y$, $h(y|c)$ can be rewritten as 
\begin{align}
    h(y|c) = \left(\sum_{x\in \Omega} {h(x|c)\over h(y|c)}\right)^{-1} = \left(\sum_{x\in \Omega} r(x,y|c) \right)^{-1},
\end{align}
where $\Omega$ is the full vocabulary. That’s better, but that would still cost around 50,000 questions (the number of tokens) for each value of $h(x|c)$. To lower this number we use importance sampling. The bulk of $\sum_x r(x,y|c)$ is where the most likely tokens are, so it’s not worth asking for every one of them. We condition a reference language model $G$ on our context $c$ from which we can sample the most likely tokens. We can then estimate $h(y|c)$ with the following approximation:
\begin{equation}
    h(y|c) = \left( \mathbbm{E}_{X\sim G_c}\left[{r(X,y|c) \over g(X|c)}\right] \right)^{-1} \approx \left( {1\over n}\sum_{x\sim G(c)} {r(x,y|c) \over g(x|c)} \right)^{-1},
\end{equation}
where $n$ is the number of tokens generated by $G$, which can be much smaller than 50,000. $g(x|c)$ is the probability of $x$ given $c$ outputted by $G$. $\sum_{x\sim G(c)}$ is the sum over the set of tokens that the model $G$ conditioned on $c$ generates in practice when queried $n$ times. So in each round of the language model game, one candidate token $x$ is the true next token, while the other, $y$, is generated by $G(c)$.

To decrease the variance of this estimator for $h(y|c)$, we instead estimate
\begin{equation}
{h(y|c)\over g(y|c)} = \left( \mathbbm{E}_{X\sim G_c}\left[{g(y|c) \over g(X|c)} r(X,y|c) \right] \right)^{-1}\approx \left( {1\over n}\sum_{x\sim G(c)} {g(y|c)\over g(x|c)} r(x,y|c) \right)^{-1}.
\label{eq:human_distro_estimator}
\end{equation}
The variance is lower because, if $h$ and $g$ are close, most terms in the sum will be close to 1 (whereas $r(x,y|c)/g(x|c)$ can get very large or small on some samples).

With access to $h(y|c)/g(y|c)$, we can compute $L_h - L_G$ (where $L_G$ is the loss of the reference language model $G$ defined by equation \ref{eq:human_loss_def}, with $h$ replaced with $g$). Using $N$ samples from the true target corpus, from equations \ref{eq:human_loss_def} and \ref{eq:human_distro_estimator}, we find
\begin{align}
\label{eq:loss_estimation}
    L_h-L_G = \mathbbm{E}_{(C,Y)\sim T}\left[\log \frac {g(Y|C)} {h(Y|C)}\right] \approx {1\over N} \sum_{(c,y)\sim T} \log\left( {1\over n} \sum_{x\sim G_c} {g(y|c) \over g(x|c)} r(x,y|c) \right).
\end{align}
We can then add the known value for $L_G$ to find $L_h$, and from that, $P_h$.
In practice, we use GPT-2-small (117M parameters) as our generator language model $G$.

\subsubsection{Controlling for sample bias in importance sampling}
\label{sec:sample_bias}

Due to the fact that one of the two candidate tokens is sampled from a language model $G$, the participants no longer have an incentive to submit their actual guess at $r(x,y|c)$, since they have extra information about how $x$ and $y$ are sampled.

Naively, we would find $r(x,y|c) = h(x|c)/h(y|c)$ by making the human guess if the prompt $c$ is followed by $x$ or $y$, and then the human should answer that $c$ is followed by $x$ with probability $h(x|c)/(h(x|c)+h(y|c))$. This would be true if one token was sampled from the true distribution and the other one was selected uniformly among all other tokens. However, this isn't true here because the other token is sampled from $G_c$. 

A rational agent that perfectly knows $T$ and $G$ would answer that $c$ is followed by $x$ with probability
\begin{equation}
    {P(x\sim T(c) \wedge y\sim G(c)) \over P(x\sim T(c)\wedge y\sim G(c)) + P(x\sim G(c) \wedge y \sim T(c))} \\ = {t(x|c) g(y|c) \over t(y|c)g(x|c) + t(x|c) g(y|c)},
\end{equation}
where $x~G(c)$ refers to the event "$x$ was sampled from the distribution $G(c)$", $t(x|c)$ is the true distribution $T$. In the second line, we used the independence of $T$ \& $G$. A human could use their knowledge of $G$ as extra information that can help guess the right answer.
For example, a human who believes $T(c)$ to be indistinguishable from $G(c)$ will answer 0.5 to every question, making it impossible to extract $h(x|c)/h(y|c)$.

The solution is to incentivize the human to give something other than their best guess. We ask the human for the probability $p$ that $c$ is followed by $x$, and we reward them with a weighted binary cross-entropy reward
\begin{align}
    R(p) = g(x|c) \log (p) 1_{z=x} + g(y|c) \log (1 - p) 1_{z=y}.
\end{align}
where $z$ is the correct answer. The expected value of this reward, according to a human believing that the generative model follows a distribution $\hat{G}$, is
\begin{equation}
    \mathbbm{E}[R(p)]=\hat{g}(x|c)\log(p)(h(x|c)\hat{g}(y|c))+\hat{g}(y|c)\log(1 - p)(h(y|c)\hat{g}(x|c)).
\end{equation} 
This is at its maximum when $d\mathbbm{E}[R(p)]/dp=0$, therefore, the optimal play satisfies
\begin{equation}
    {\hat{g}(x|c)\hat{g}(y|c)h(x|c) \over p^{\ast}} = {\hat{g}(y|c) \hat{g}(x|c) h(y|c) \over 1 - p^{\ast}}
    \Rightarrow p^* = {h(x|c) \over h(x|c) + h(y|c)}.
\end{equation}
If we assume that humans play optimally (which is a questionable assumption), then the human's response, $p^*$,  represents the human's true belief for $h(x|c)/(h(x|c)+h(y|c))$, no matter what their beliefs about the generative model are.

In practice, we use a slightly different reward, $R(p)=1000(g(x|c)(\log(p) - \log(0.5))1_{z=x} +g(y|c)(\log(1 - p) - \log(0.5))1_{z=y}$, for which the optimal play is the same, but is more understandable for a human. Participants get a reward of 0 for saying $p=0.5$, and the scaling makes their score more readable.

On rounds where $x=y$, the participant isn't asked to compare their relative likelihoods. Instead, the website automatically answers that both are as likely.

\subsubsection{Estimating uncertainty in our measurement of perplexity}
\label{sec:uncertainties}

If we assume humans play rationally in the game described above, there remain three sources of uncertainty in our measurement of human perplexity:
\begin{enumerate}[leftmargin=.9cm]
    \item 
    {\bf{Uncertainty in the sum over $(c,y)\sim T$.}} We collected data on only 120 different $(c,y)$ pairs, which is small considering that the per-token loss  has a large variance. This causes a non-negligible uncertainty over the measured perplexity, even if we had perfect estimates of each value of $h(y|c)/g(y|c)$ for each $(c,y)$ pair. We compute the empirical standard deviation $\sigma$ on $L_h$ over $(c,y)$ pairs. This gives us a lower bound $\exp(L_h-2\sigma)$ and an upper bound $\exp(L_h+2\sigma)$ on the perplexity $P_h=\exp(L_h)$.
    \item
    {\bf{Uncertainty in the sum over $x\sim G(c)$.}} We now consider the uncertainty in each $h(y|c)/g(y|c)$ sample estimated by equation \ref{eq:human_distro_estimator}. A natural way to estimate this is to use the empirical standard deviation over the sum in equation \ref{eq:human_distro_estimator} and propagating this uncertainty to $L_h$. But the sum in equation \ref{eq:human_distro_estimator} is heavy-tailed. This means that, since we only have a small amount of samples ($\approx 10$) for every prompt, we underestimate the perplexity of any predictor using this technique. The degree of underestimation cannot be easily quantified, as the weight of the heavy-tail examples depends on the distance between human predictions and the reference LM prediction, as can be seen in Figure \ref{fig:perplexity}. To test this effect, we estimated the loss of a 14M-parameter LM and a 345M-parameter LM, using the 117M-parameter LM as a generator. We find the underestimation to be at most 0.5 bits away from the ground truth when $n\approx 40$, for models that are very dissimilar (like a 14M-parameter model vs a 117M-parameter LM, for which the difference in true loss is 1.3 bits). The results are shown in Figure \ref{fig:loss_underestimation}. This is a large difference, but this is still good enough for the purpose of comparing humans to language models because we find differences between humans and LMs much larger than $\sim 0.5$ bits. Hence we do not attempt to estimate this error, and accept it as a limitation of our study.
    
\begin{figure}[h]
    \centering
    \vspace{-12mm}
    \includegraphics[width=0.6\textwidth]{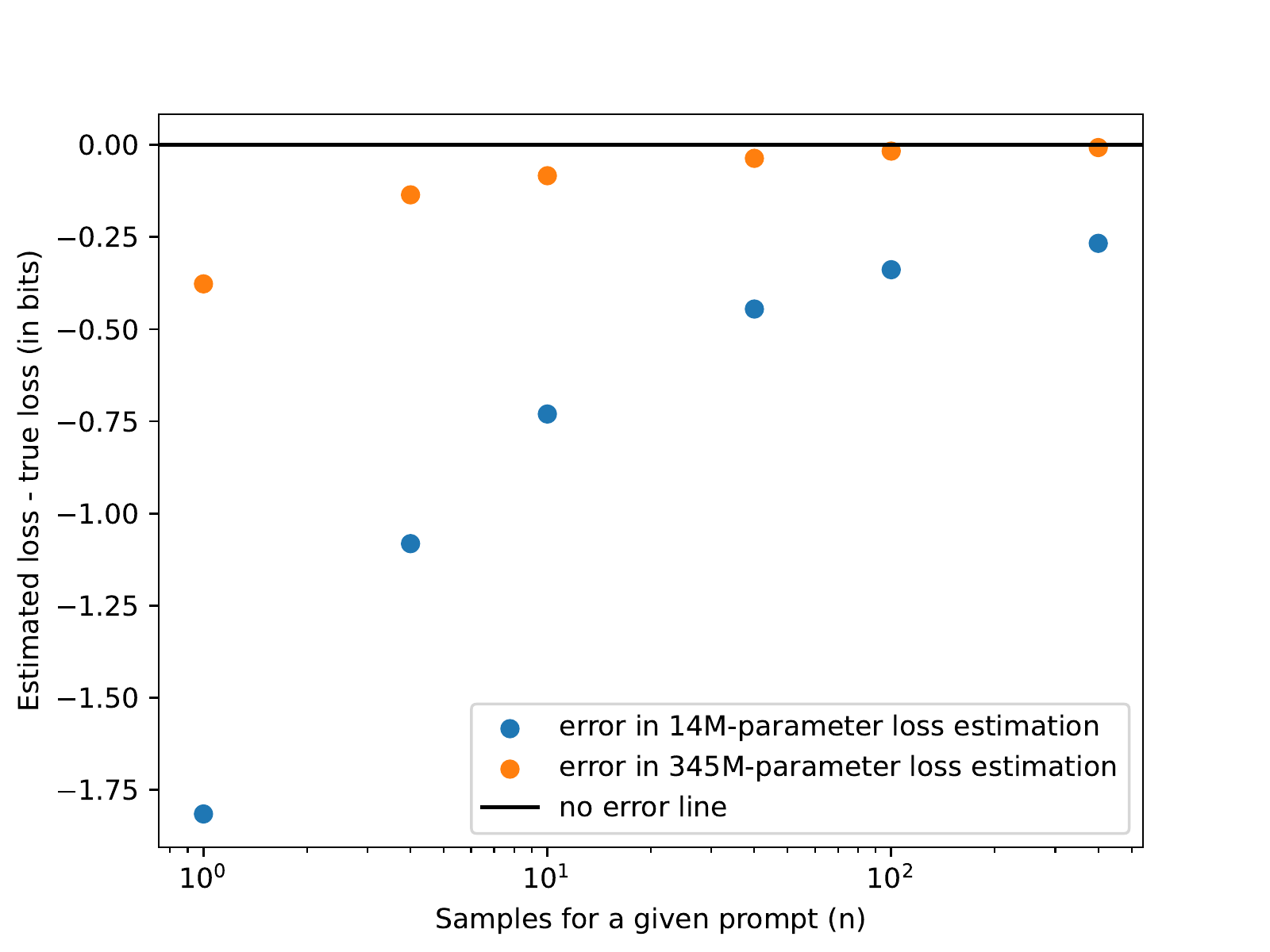}
    \caption{Difference between the loss estimated by equation \ref{eq:loss_estimation} and the ground truth loss for two LMs, for varying $n$ (number of samples per prompt), for $N=1000$ prompts.}
    \label{fig:loss_underestimation}
\end{figure}
\end{enumerate}
We only include 1. in our final uncertainties (shown as error bars in Figure \ref{fig:perplexity}). Measuring uncertainty by bootstrapping (rejecting random halves of users) yields similar results (see Appendix \ref{sec:boostrap}).

\subsubsection{Estimating perplexity for language models}

To ensure a true apples-to-apples comparison, 
we estimated the perplexity of the language models in the same way as described for humans above, on the same dataset. We used GPT-2 models - GPT-2 medium (345M parameters) and small (117M parameters) \citep{radford2019language}. We also used a 2-layer 14M-parameter model we trained ourselves on the train set of OpenWebText as an example of a minimal model.

As our human participants could only enter one of the 11 ratios in our interface (99\%, 90\%, 80\%, 70\%, 60\%, 50\%, 40\%, 30\%, 20\%, 10\%, or 1\%), we also report the “rounded” performance of our LMs - that is, the performance of our LMs if they choose the checkbox that is the closest to their probability ratio. (Note that we can’t access the true perplexity of rounded models, as only ratios are rounded and not the probability of the correct token.). As shown in Appendix \ref{sec:rounding}, more extreme rounding leads to much worse performance, and humans would need to produce calibrated and extreme probabilities to get low performance. 

\vspace{-2mm}
\subsection{Results and limitations}

\vspace{-2mm}
\begin{figure}[!b]
    \centering
    \vspace{-4mm}
    \includegraphics[width=1\textwidth]{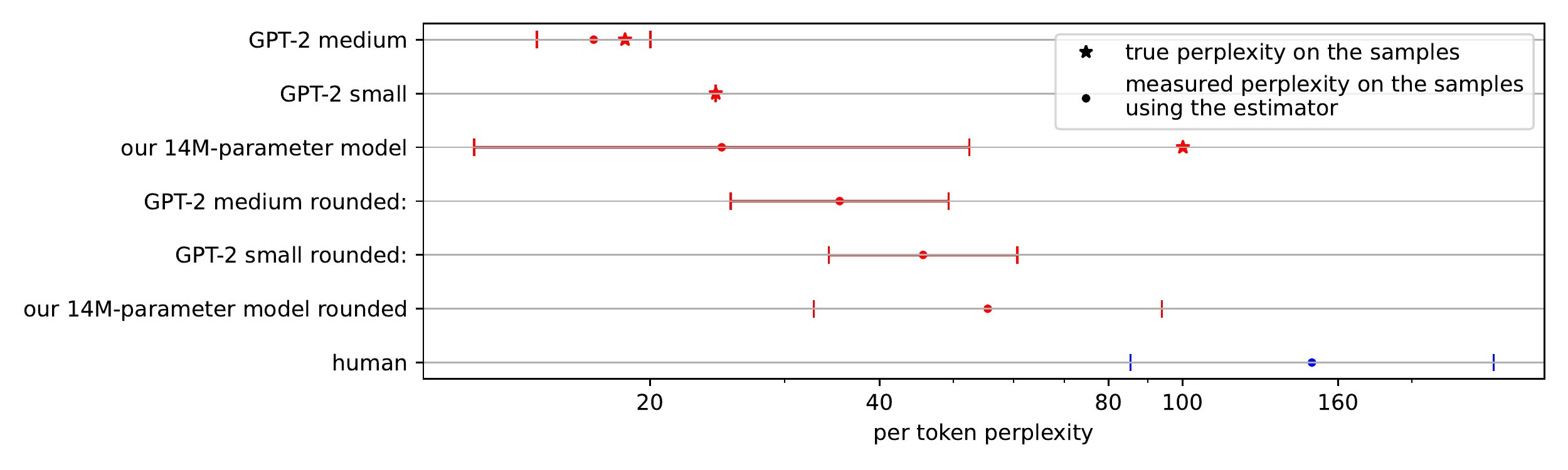}
    \vspace{-4mm}
    \caption{Our estimated perplexities for a number of language models and humans. The human perplexity was obtained from our study described in Section \ref{sec:perplexity_method}. GPT-2 small has no estimated value, since this was used as the reference generator model. Error bars are determined according to uncertainty source (i) described in Section \ref{sec:uncertainties}.}
    \label{fig:perplexity}
\end{figure}
Our estimates of perplexity for humans and LMs, along with the uncertainty estimate, are shown in Figure \ref{fig:perplexity}. Since we used the same method for estimating perplexity for humans and LMs, this is a direct comparison between human and LM performance. As explained in Section \ref{sec:uncertainties}, all losses obtained by this method are underestimations of the true loss. GPT-2-small is used as the generator, which is why its measured perplexity using the estimator is perfect.

This method could also overestimate human perplexity: there could be other setups in which it would be easier for players to give calibrated probabilities. In fact, some players found the scoring system hard to understand, and if it led them to not express their true probability ratios, we might have underestimated human performance. In general, this method is very sensitive to failures at giving calibrated probability estimates: the high perplexity obtained here is probably partially due to humans being bad at giving calibrated probabilities, rather than humans just being bad at language modeling. In addition, as humans are restricted to one of 11 ratios, our setup could also underestimate performance by artificially reducing the resolution of our human participants.

Finally, this evaluation also shares the weaknesses of our Top-1 accuracy measurements: humans could struggle with tokenization in a way that would be solved by more training, and our reference language models may be overfitted to the evaluation dataset. We can't reproduce the analysis done with GPT-Neo models since the training set of GPT-2 models is unknown, but given that the perplexity values reported in Figure \ref{fig:perplexity} are close to the validation perplexity of Table \ref{table:acc-and-perplexity} (and are measured on similar datasets using the same tokenizer), it is unlikely that our LLM perplexity measurements are far from their true value. We also find that excluding non-word tokens from our analysis does not change qualitative takeaways of our experiments (see Appendix \ref{sec:exclude}).

Thus, while we don’t have a good way to precisely measure human perplexity, these results give reasonable evidence that it is high. In particular, humans are worse than a GPT-2-small at giving calibrated estimates of token vs token probabilities. 

\section{Discussion}
\label{sec:conclusion}

The results here suggest that humans are worse than even small language models the size of GPT-2-small (117M parameters) at next-token prediction, even on the top-1 prediction task. This is true even when the humans have practiced for 1-2 hours. Some humans may beat medium-sized language models (GPT-2-small, GPT-Neo-1.3B) with enough practice, but it seems very unlikely that even the best humans could beat substantially larger models.

These results may be surprising because humans are better at writing coherent text than GPT-2 and therefore one might expect humans to be better at next-token prediction. But actually, these tasks are very different – if you train an autoregressive model to imitate human text, the model has to dedicate capacity to all the different features that might be informative for guessing the next token (including features that don’t affect human judgments of coherence). Language models are trained to guess what are all plausible tokens that could follow, while being coherent only requires finding one reasonable completion. Additionally, predicting the next token on OpenWebText sometimes requires very broad knowledge that may require an amount of memory that may exceed human limits.

These findings are important for language model interpretability work, because it means you should expect a language model to have more knowledge of the text than you have. It has been argued in \citet{olah2019} that models may become more interpretable as they get to human level, and then become less interpretable again as they become superhuman. The fact that existing LMs are already superhuman (at the task they’re trained on) is worth bearing in mind here.

\section{Conclusion}

In this paper, we compared the capabilities of humans and language models (LMs) in next-token prediction tasks, using top-1 accuracy and perplexity as our metrics. We find that even relatively small language models outperform humans at next-token prediction.
\vspace{-2mm}

\section*{Acknowledgements and contributions}
\vspace{-2mm}

We thank a variety of people for comments and assistance (especially Paul Christiano, Nostalgebraist, and Rafe Kennedy), and to various people for playing the game. The top-1 prediction web app was mostly written by Buck Shlegeris, Fabien Roger designed and ran the perplexity experiment and performed most of the analysis, Lawrence Chan performed the research on previous measurements, and Euan McLean adapted the Alignment Forum post into a paper. Thanks to Paul Christiano for suggesting the rough approach for our human perplexity experiment, and to Paul and Adam Scherlis for helping with some of the details.

\bibliography{main}
\bibliographystyle{tmlr}

\appendix
\section{Data collection interface}

\begin{figure}[H]
    \centering
    \vspace{-6mm}
    \includegraphics[width=0.7\textwidth]{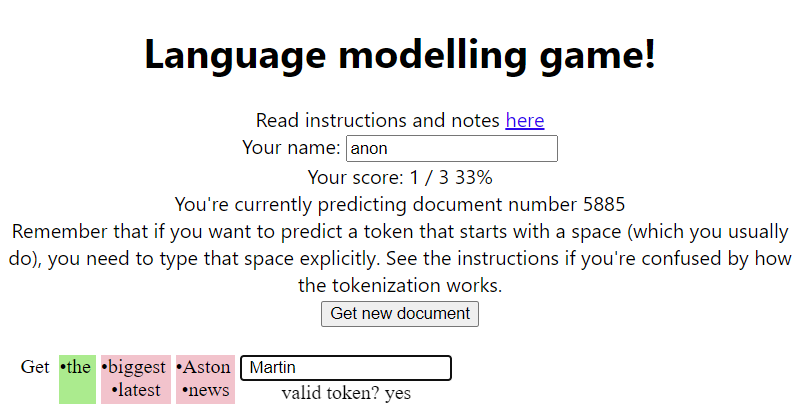}
    \caption{Interface for the experiment described in Section \ref{sec:human_top1accuracy}. The interface is available at \url{https://rr-lm-game.herokuapp.com}}
    \label{fig:top1game}
\end{figure}

\begin{figure}[H]
    \centering
    \vspace{-5mm}
    \includegraphics[width=1\textwidth]{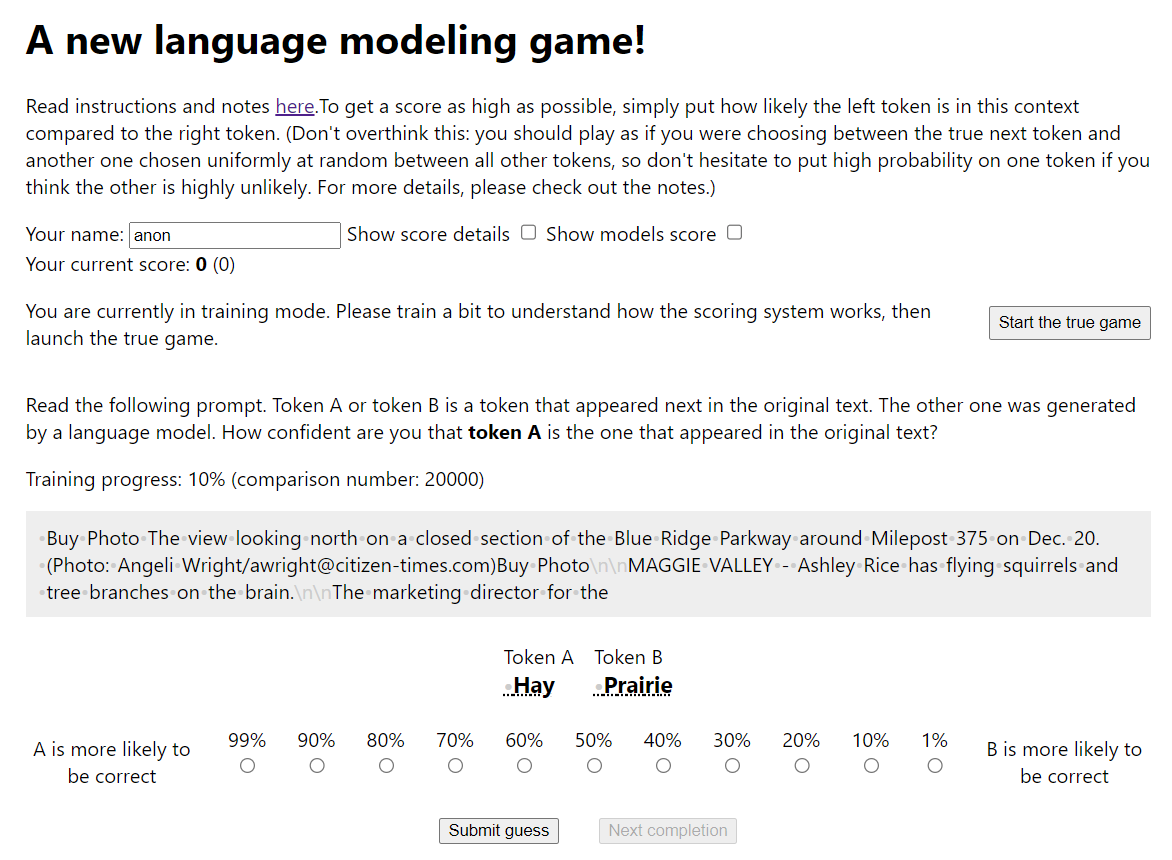}
    \vspace{-2mm}
    \caption{Interface for the experiment described in Section \ref{sec:human_perplexity}. The interface is available at \url{https://rr-lm-game.herokuapp.com/whichonescored}.}
    \label{fig:perplexitygame}
\end{figure}

Figure \ref{fig:top1game} and \ref{fig:perplexitygame} show the interfaces participants used.

\section{Uncertainty estimation using bootstrapping} \label{sec:boostrap}

\begin{figure}[H]
    \centering
    \includegraphics[width=0.9\textwidth]{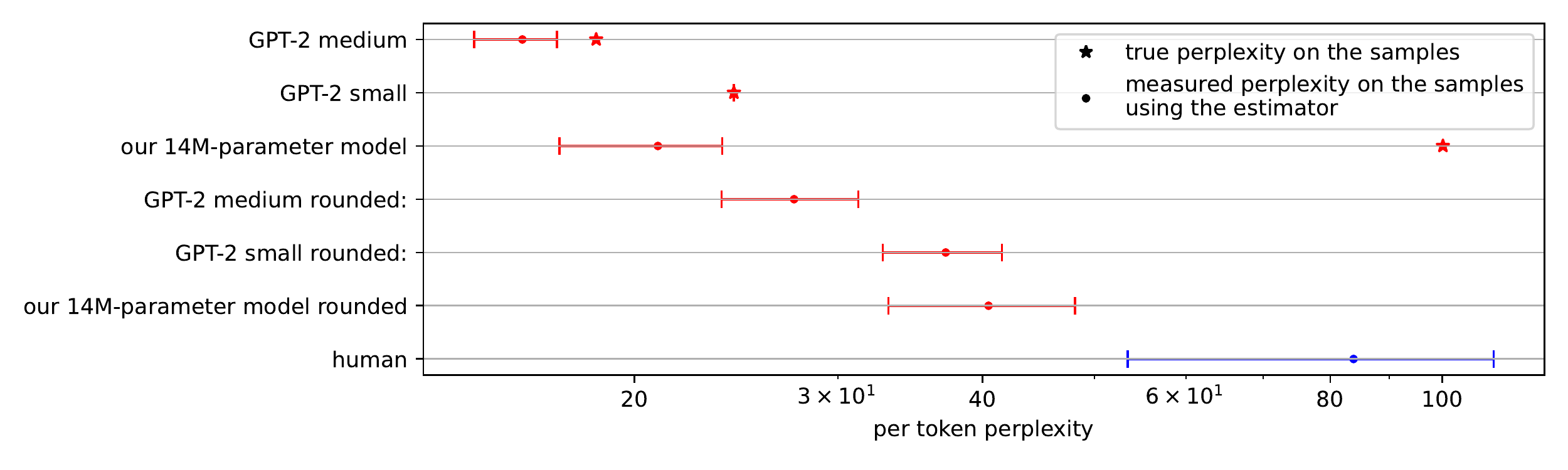}
    \vspace{-2mm}
    \caption{Our estimated perplexities for a number of language models and humans when using bootstrapping to estimate uncertainty. Error bars show the median estimate, as well as 0.05 and 0.95 quantiles. }
    \label{fig:boostrap}
\end{figure}

Figure \ref{fig:boostrap} shows perplexity when estimating uncertainties with bootstrapping: we compute perplexity 100 times, and ignore the entries of a random half of users on each iteration. We report the 0.05 and 0.95 quantiles as uncertainty estimates. We find that this reduces the estimate of perplexity for both humans and language models, which is not surprising given that our estimate of perplexity underestimates the true perplexity more when the number of samples is smaller (as shown in Figure \ref{fig:loss_underestimation}). Human perplexity remains above language models', which shows that our conclusion does not depend on the poor performance of a few users.

\section{Influence of Rounding on Perplexity} \label{sec:rounding}

\begin{figure}[H]
    \centering
    \includegraphics[width=0.9\textwidth]{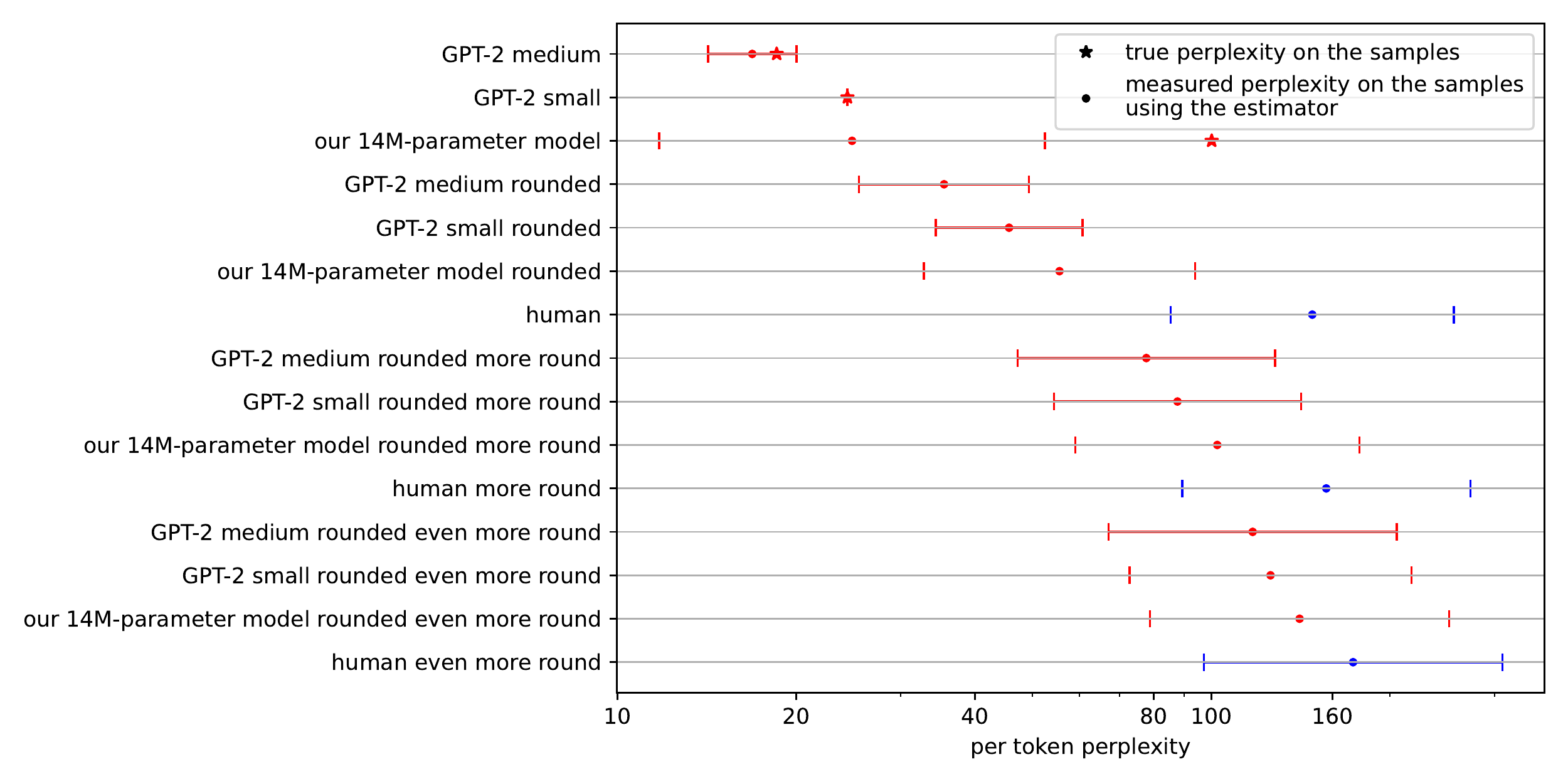}
    \caption{Our estimated perplexities for a number of language models and humans with more extreme rounding. "rounded" corresponds to the level of rounding of the interface, "more round" corresponds to rounding to the nearest of [0.1, 0.2, 0.5, 0.8, 0.9] (human answers are rounded post-hoc) and "even more round" corresponds to rounding to the nearest of [0.2, 0.5, 0.8]. }
    \label{fig:round}
\end{figure}

Figure \ref{fig:round} shows the perplexity when more rounding is applied.  We find that applying more rounding significantly increases language model perplexity without increasing human perplexity much, which shows the importance of using the relatively fine-grained rounding we used.

\section{Perplexity Estimation on Word-tokens} \label{sec:exclude}

\begin{figure}[H]
    \centering
    \includegraphics[width=0.9\textwidth]{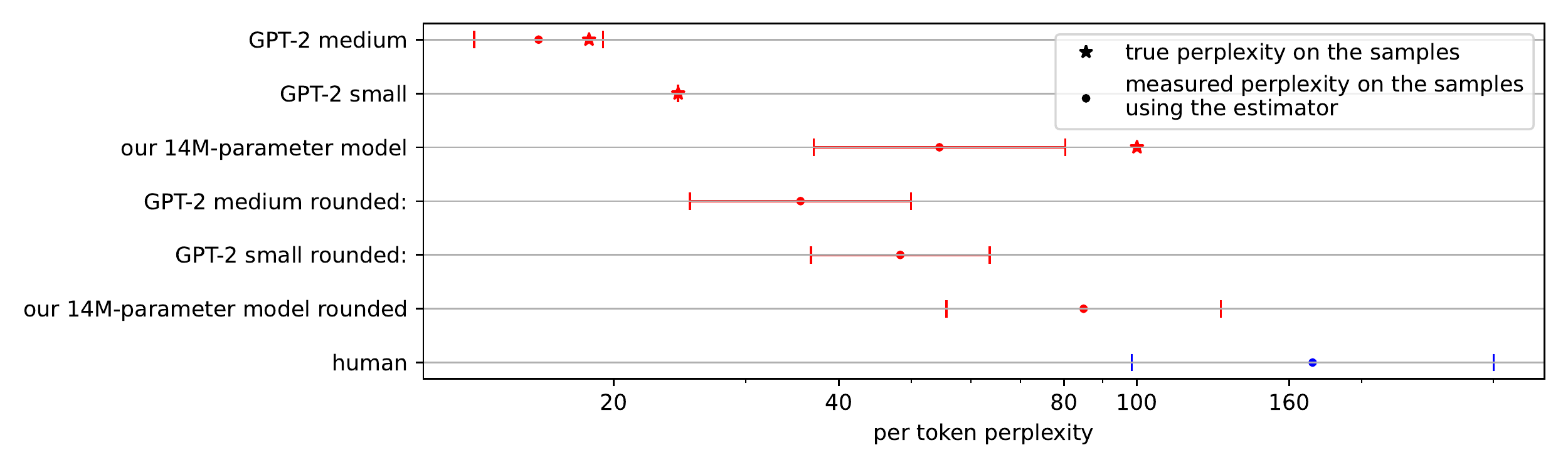}
    \caption{Our estimated perplexities for a number of language models and humans when only using questions about word tokens. }
    \label{fig:filtered}
\end{figure}

Figure \ref{fig:filtered} shows perplexity estimates when only keeping the 84/120 choices where the correct token is a word (a token that starts with a space and otherwise only contains alphabetical characters). This does not change the conclusions drawn from the main experiments.

\end{document}